\pdfoutput=1

\documentclass[11pt]{article}

\usepackage[preprint]{acl}

\usepackage{times}
\usepackage{latexsym}

\usepackage[T1]{fontenc}

\usepackage[utf8]{inputenc}

\usepackage{microtype}

\usepackage{inconsolata}

\usepackage{amsmath, amsfonts, amssymb, amsthm}

\usepackage{graphicx}

\title{Unraveling the Localized Latents: Learning Stratified Manifold Structures in LLM Embedding Space with Sparse Mixture-of-Experts}

\author{Xin Li \\
  Rutgers, the State University \\of New Jersey, Piscataway, NJ, USA \\
  \texttt{xl598@rutgers.edu} \\\And
  Anand D. Sarwate \\
  Rutgers, the State University \\of New Jersey, Piscataway, NJ, USA \\
  \texttt{anand.sarwate@rutgers.edu} \\}

\begin{document}
\maketitle
\begin{abstract}
Large Language Models (LLMs) produce rich, high-dimensional embeddings that capture semantic relationships in the input space. However, real-world data often exhibit complex local structures that can be challenging for single-model approaches with a smooth global manifold in the embedding space to unravel. In this work, we conjecture that in the latent space of these large language models, the embeddings live in a local manifold structure with different dimensions depending on the perplexities and domains of the input data—commonly referred to as a Stratified Manifold structure, which in combination form a structured space known as a Stratified Space. To investigate the validity of this structural claim, we propose an analysis framework based on a Mixture-of-Experts (MoE) model where each expert is implemented with a simple dictionary learning algorithm at varying sparsity levels. By incorporating an attention-based soft-gating network, we verify that our model learns specialized sub-manifolds for an ensemble of input data sources, reflecting the semantic stratification in LLM embedding space. We further analyze the intrinsic dimensions of these stratified sub-manifolds and present extensive statistics on expert assignments, gating entropy, and inter-expert distances. Our experimental results demonstrate that our method not only validates the claim of a stratified manifold structure in the LLM embedding space, but also provides interpretable clusters that align with the intrinsic semantic variations of the input data.
\end{abstract}

\section{Introduction}

Large Language Models (LLMs) such as GPT \cite{gpt, gpt4}, Llama \cite{llama, llama2, llama3}, and Claude \cite{claude} generate rich embeddings that have demonstrated state-of-the-art performance in numerous tasks in Natural Language Processing (NLP) \cite{llm-strats}. Nevertheless, these embeddings reside in high-dimensional spaces where data often exhibit non-uniform, complex local structures. We hypothesize that LLM embeddings do not lie on a single global manifold but rather form a \textit{Stratified Space} \cite{stratified-morse-theory}, a union of lower-dimensional local manifolds (or strata) of different dimensions that could correspond to data of different domains, perplexities (entropy), or semantic concepts, \textit{etc}.

In this paper, we introduce a Dictionary Learning-based Mixture-of-Experts (MoE) framework designed to examine this stratified structure of the embedding spaces in large language models, where each dictionary learning expert has its own sparsity level to get a better idea of expert assignments in different strata. These experts are then combined with an attention-based soft-gating network to dynamically route each input data sample to the most suitable expert(s), which implicitly leads to a favorable choice regarding the dimension of the local sub-manifold (strata), effectively partitioning the embedding space into specialized sub-manifolds (or \textit{Stratified Manifolds}).

To summarize, in this paper we lay out a basic framework to validate our stratified manifold hypothesis in the context of LLM embedding spaces, which includes,
\begin{enumerate}
    \item A complete MoE-based pipeline to analyze the expert assignments for each input data sample, which sheds light on the intrinsic structure of the strata in which the input data might live in the embedding space.
    \item Empirical evaluations of the embedding stratification, which can also provide some insights into the interpretability and performance of the large language models.
    \item Visualizations to gain a better understanding of the LLM working mechanisms in the context of stratifided spaces.
\end{enumerate}

\section{Related Work}

\begin{description}
    \item[Geometry of LLM Embeddings] Previous studies on the geometric properties of LLM Embeddings suggest that the embeddings capture extensive semantic information but are often distributed anisotropically, which can lead to high similarity between unrelated words, limiting the expressive power of the model \cite{anisotropic}. \citeauthor{token-strats} have shown that the LLM token embedding space indeed exhibits a stratified manifold structure.
    \item[Mixture-of-Experts] Originally introduced by \citeauthor{moe}, the technique uses a gating network that routes inputs to one or a sparse set of specialized expert models. In recent years, MoE has scaled up drastically to work with large language models to increase model capacity with minimal added computations \cite{moe-llm}. Our approach also uses an MoE architecture, but for a different purpose: instead of scaling capacity, we use MoE to cluster and capture structure in an existing embedding space with known expert hyperparameters (such as sparsity level in a dictionary learning algorithm). In essence, we employ an unsupervised MoE that operates on fixed LLM embeddings to find meaningful partitions of different geometrical structures.
    \item[Dictionary Learning \& Sparse Coding] Dictionary learning is an algorithm that learns a set of dictionary atoms that can sparsely represent data through a linear combination (sparse codes) of atoms \cite{dl}. Various algorithms have been proposed over the years that can learn dictionary and sparse codes effectively include KSVD \cite{ksvd, ksvd-efficient}, Online Dictionary Learning \cite{odl}, Orthogonal Matching Pursuit (OMP) and its variant Batch-OMP \cite{omp, ksvd-efficient}, Iterative Hard-Thresholding (IHT) and Iterative Soft-Thresholding Algorithm (ISTA) \cite{cs}, Learnable-ISTA \cite{lista}, \textit{etc}.
\end{description}

\section{Preliminaries}

In this section, we provide a basic overview of the relevant concepts to help understand our work.

\subsection{Mathematical Foundations of Stratification}

Here, we provide a quick review of some of the core mathematical concepts and methodologies to help understand the idea of stratification presented in this paper. However, these concepts do require a basic understanding of mathematical analysis \cite{analysis}, topology \cite{topology} and algebra \cite{abstract-algebra, commutative-algebra}, for a more comprehensive view of the other basic mathematical concepts discussed in this paper, please refer to Appendix \ref{subsec:appendix-review}.

\begin{definition}[Topological Equivalence or Homeomorphism]
    Let $\mathsf{X}$ and $\mathsf{Y}$ be two topological spaces. A function $f: \mathsf{X} \mapsto \mathsf{Y}$ is \textit{continuous} if for each point $x \in \mathsf{X}$ and each neighborhood $\mathsf{N}$ of $f(x)$ in $\mathsf{Y}$ the set $f^{-1} (\mathsf{N})$ is a neighborhood of $x \in \mathsf{X}$. A function $h: \mathsf{X} \mapsto \mathsf{Y}$ is called a homeomorphism if it is one-to-one, continuous, and has a continuous inverse function. When such a function exits, $\mathsf{X}$ and $\mathsf{Y}$ are called \textit{homeomorphic} (or \textit{topologically equivalent}) spaces.
\end{definition}

A space whose topology has a countable base is called a \textit{second countable} space, which loosely means to limit how ``large'' a space can be and ensures that the space is separable.

\begin{definition}[Hausdorff Space]
    A topological space with the property that two distinct points can always be surrounded by disjoint open sets is called a \textit{Hausdorff space}.
\end{definition}

Essentially, Hausdorff spaces are the spaces where any two points being ``far off'' is concretely defined. With these definitions in mind, we can proceed to provide a formal definition of manifolds,

\begin{definition}[Manifold]
    A \textit{manifold} of dimension $n$ ($n$-manifold) is a second-countable Hausdorff space, each point of which has a neighborhood homeomorphic to the Euclidean space $\mathbb{R}^n$.
\end{definition}

From the above definitions, we can loosely understand the manifolds as topological spaces (space preserving locality but invariant to distance metrics) that look like an Euclidean space locally.

With these concepts in mind, we are ready to define stratification,

\begin{definition}[Stratified Spaces]
    A \textit{stratification} of a topological space $\mathsf{X}$ is a filtration of the decomposition $\mathsf{X} = \bigcup_{i = 0}^n \mathsf{S}_i$,
    \begin{equation}
        \emptyset = \mathsf{S}_{-1} \subset \mathsf{S}_0 \subset \mathsf{S}_1 \subset \cdots \subset \mathsf{S}_{n - 1} \subset \mathsf{S}_n = \mathsf{X},
    \end{equation}
    where each of the $\mathsf{S}_i$'s is a smooth manifold, or \textit{stratified manifold} (possibly empty) of dimension $i$ and hence
    \begin{equation}
        \mathsf{cl}(\mathsf{S}_k) \setminus \mathsf{S}_k \subset \bigcup_{i = 0}^{k - 1} \mathsf{S}_i,
    \end{equation}
    the closure $\mathsf{cl}(\mathsf{S}_k)$ is called the stratum of dimension $k$.
\end{definition}

\begin{figure*}[hbt!]
    \centering
    \includegraphics[width=\textwidth]{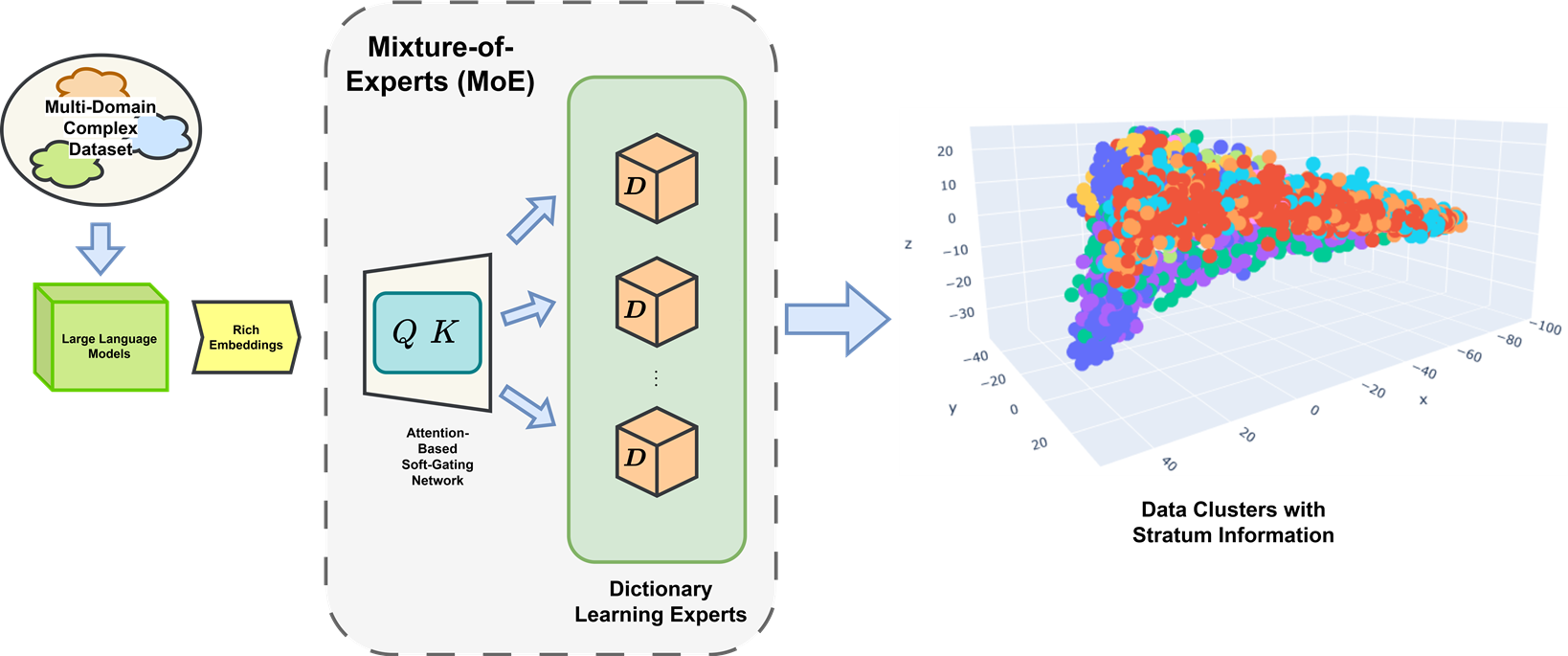}
    \caption{Working mechanism of our model to learn stratum information from a complex dataset.}
    \label{fig:moe}
\end{figure*}

\subsection{Mixture of Experts (MoE): An Introduction}

Mixture of Experts (MoE) is a modeling approach where an ensemble of specialized models, namely, the ``experts'', are combined through a gating network to solve a complex task \cite{moe, moe-llm, moe-bigdata}. This framework allows the model to divide the input space into regions and assign different experts to different regions, effectively partitioning the problem into simpler sub-problems that each expert can handle well, and can also provide us some dimensional information of the learned sub-regions based on the predefined complexity of each expert.

Formally, an MoE model consists of the following components:
\begin{itemize}
    \item \textbf{Expert Networks}: a modern deep learning based MoE model consists of $E$ expert networks defining functions $f_1, f_2, \dots, f_E$, each will take the same input $x$ and produce an output $f_e(x)$ (\textit{e.g.}, some form of predictions or transformed representations).
    \item \textbf{A Gating Mechanism}: a deep network based gating network $g$ will take the input $x$ and produce a set of non-negative weights,
    \begin{equation}
        \boldsymbol{w} = g(x) = \{ w_1, w_2, \cdots, w_E \},
    \end{equation}
    Typically, $\boldsymbol{w}$ is implemented as a probability distribution over expert functions,
    \begin{equation}
        \sum_{e = 1}^E w_e = 1,\: \text{ where } w_e \geq 0.
    \end{equation}
    \item \textbf{Mixture output}: Given an input $x$, the MoE model will combine the expert outputs into a single output with the gating weights. Specifically, let $y$ denote the overall output, we have,
    \begin{equation}
        y \;=\; \sum_{e=1}^E w_e\, f_e(x)\,,
    \end{equation}
    so that the gating weights $w_e$ determine how much each expert's output contributes to the final output $y$.
\end{itemize}
This formulation allows the MoE to represent complex functions by blending the behaviors of multiple simpler experts, which can also be helpful for our use case where we want to examine how experts of different complexities learn in different localities of an LLM embedding space. 

In modern deep learning settings, an MoE model is typically trained end-to-end using stochastic gradient descent based optimization methods \cite{adam, adamw} with simultaneous updates to the expert and gating network parameters.

\section{Our Approach}

To model the embedding space as a stratified space, we focus the embeddings at the last hidden states of the model outputs. Then suppose that there are $E \in \mathsf{N}^\dagger$ experts in the MoE model, each with a learnable dictionary $\boldsymbol{D}^{(e)} \in \mathbb{R}^{M \times L}$ for $e \in [E]$ \footnote{Here $[N]$ denotes a set of positive integers upper bounded by $N \in \mathbb{N}^\dagger$, \textit{i.e.}, $[N] = \{1, 2, \cdots, N\}.$}, where $M$ is the number of dictionary atoms and $L$ is the dimension of the embedding vector. For each embedding vector $\boldsymbol{z} \in \mathbb{R}^L$, we would like to find a corresponding sparse code $\boldsymbol{\gamma}^{(e)} \in \mathsf{R}^L$ such that,
\begin{equation}
    \boldsymbol{z} \approx \boldsymbol{D}^{(e)} \boldsymbol{\gamma}^{(e)}.
\end{equation}
where we use a learning based approach for the efficient sparse coding called Learnable ISTA (LISTA) \cite{lista}. To help associate dimensional information with expert assignment, we also predefine a set of sparsity levels $\boldsymbol{s}$ for the dictionary experts, such that,
\begin{align}
    &\boldsymbol{s} = \{ s^{(1)}, \cdots, s^{(E)} \},\nonumber\\
    &\qquad \mathrm{supp}(\boldsymbol{\gamma}^{(e)}) = s^{(e)},\: \forall e \in [E]
\end{align}
where $\mathrm{supp}(\cdot)$ denotes the vector support, \textit{i.e.}, the $\ell_0$ norm. To enforce this hard sparsity constraint, for each learned sparse code vector, we select the top $s^{(e)}$ entries and use a straight-through estimator to make sure the gradient flows during backpropagation \cite{straight-through-grads}. Each dictionary is designed to capture the distinguishable linear structure of each single stratum, however, we do allow for multi-expert for a stratum with an attention-based soft-gating mechanism \cite{attention}. 

The gating network we design to perform the stratum matching is based on an attention network, which essentially performs a soft dictionary lookup that outputs a probability distribution over $E$ experts \cite{pml2}.  Specifically, let $\boldsymbol{z} \in \mathbb{R}^L$ be an input embedding vector. The gating mechanism can be broken down into the following key steps.

\begin{description}
    \item[Query Projection:] we first project the input embeddings into a low-dimensional query space:
    \begin{equation}
        \boldsymbol{q} = \boldsymbol{W}_q \boldsymbol{z},
    \end{equation}
    where the learnable query projection weight matrix $\boldsymbol{W}_q \in \mathbb{R}^{Q \times L}$, and the resulting vector $\boldsymbol{q} \in \mathbb{R}^Q$ is the query vector and $Q$ is the dimension of the query space.
    \item[Learnable Keys:] we define a set of $E$ learnable key vectors,
    \begin{equation}
        \{ \boldsymbol{k}_1, \boldsymbol{k}_2, \cdots, \boldsymbol{k}_E \},
    \end{equation}
    where $\boldsymbol{k}_e \in \mathbb{R}^Q$. The combined key matrix $\boldsymbol{k}^{E \times Q}$ provides template representations of the experts to work with the query to help the model learn expert assignments.
    \item[Similarity Calculation:] to compute the similarity between the query $\boldsymbol{q}$ and the key $\boldsymbol{k}_e$ for expert $e \in [E]$, we compute the scalar dot-product,
    \begin{equation}
        \sigma_e = \frac{\boldsymbol{q} \cdot \boldsymbol{k}_e^\top}{\sqrt{Q}},
    \end{equation}
    with the scaling factor $\sqrt{Q}$ to mitigate potential large dot-product values.
\end{description}

Then with the gating scores $\sigma_e$, we pass it to a \texttt{Softmax} layer to scale them into probabilities of soft expert assignments, \textit{i.e.}, $\sum_{e} \mathsf{softmax} (g_e) = 1$. An high-level overview of our approach is in shown in Figure \ref{fig:moe}.

\begin{figure*}[hbt!]
    \centering
    \includegraphics[width=\linewidth]{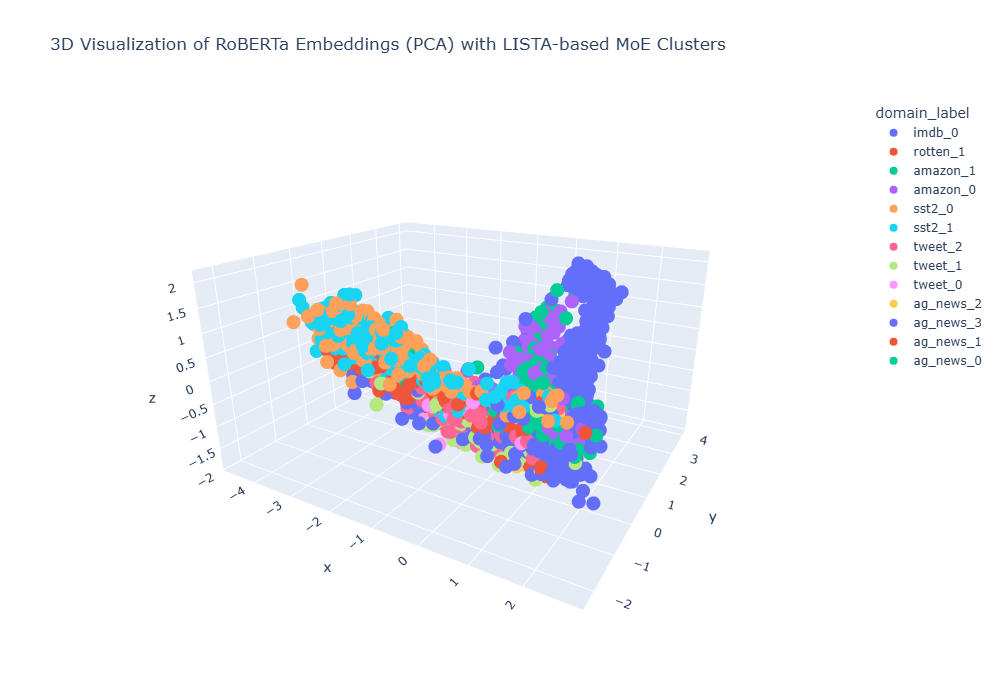}
    \label{fig:roberta}
\end{figure*}

\begin{figure*}[hbt!]
    \centering
    \includegraphics[width=\linewidth]{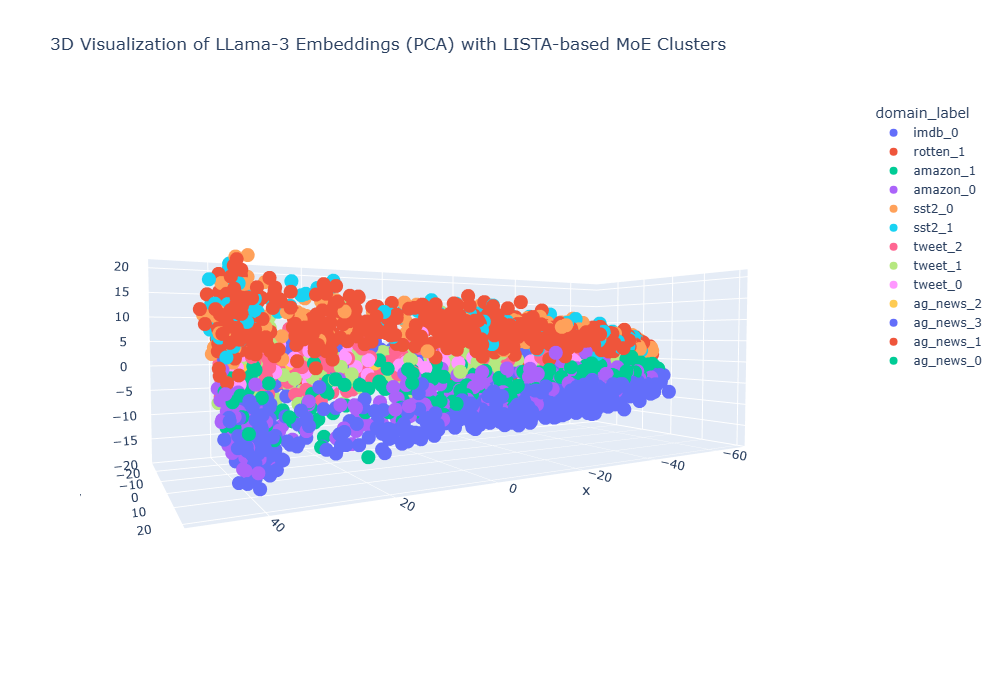}
    \label{fig:llama}
\end{figure*}

\begin{figure*}[hbt!]
    \centering
    \includegraphics[width=\linewidth]{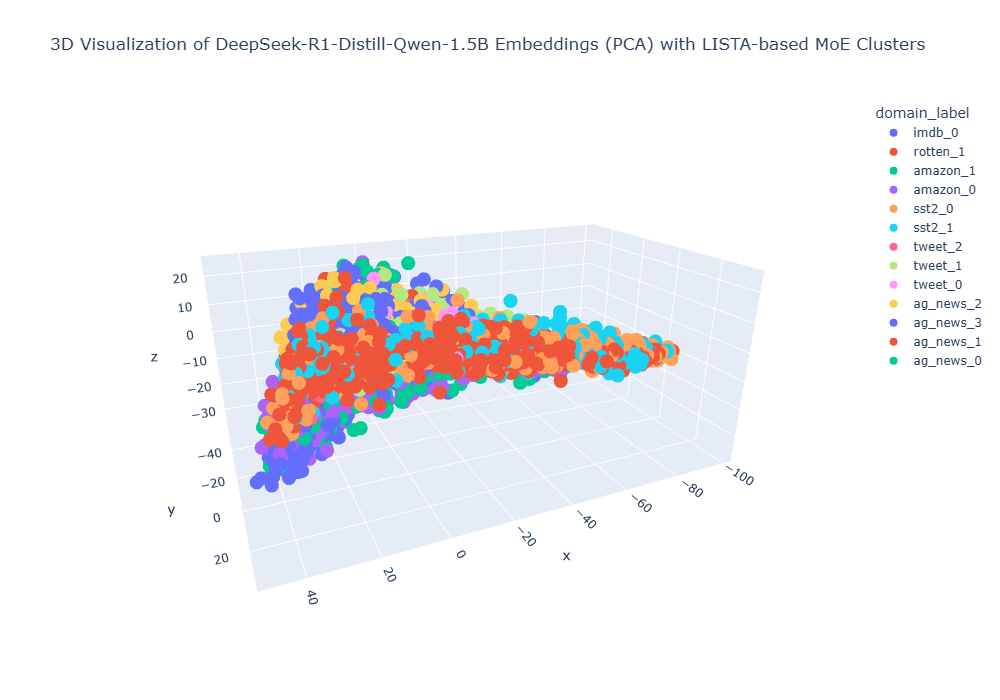}
    \label{fig:deepseek}
\end{figure*}

\begin{figure*}[hbt!]
    \centering
    \includegraphics[width=0.85\linewidth, height=12.5em]{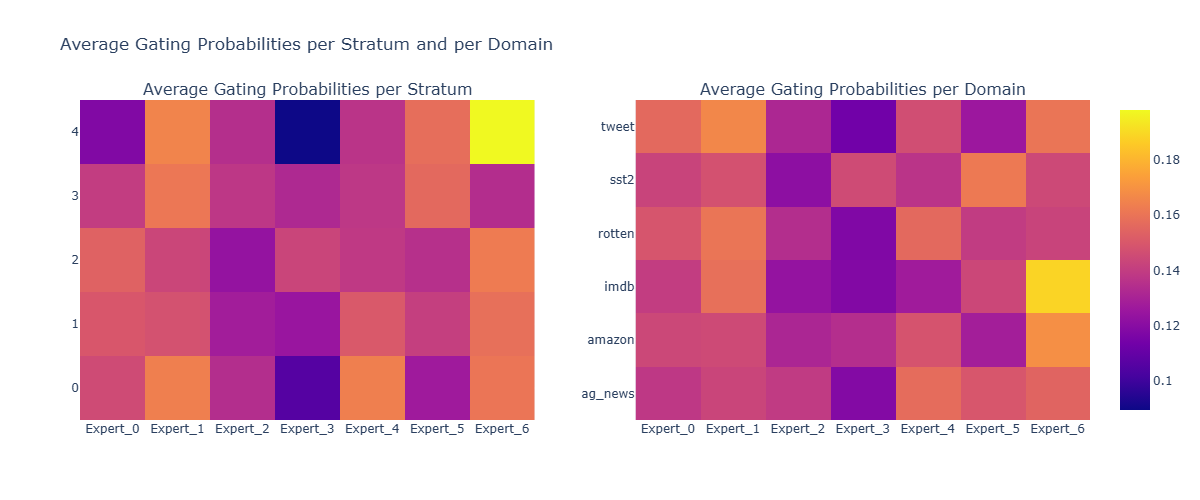}
    \caption{Gating probability visualizations for RoBERTa.}
    \label{fig:roberta-probs}
\end{figure*}

\begin{figure*}[hbt!]
    \centering
    \includegraphics[width=0.85\linewidth, height=12.5em]{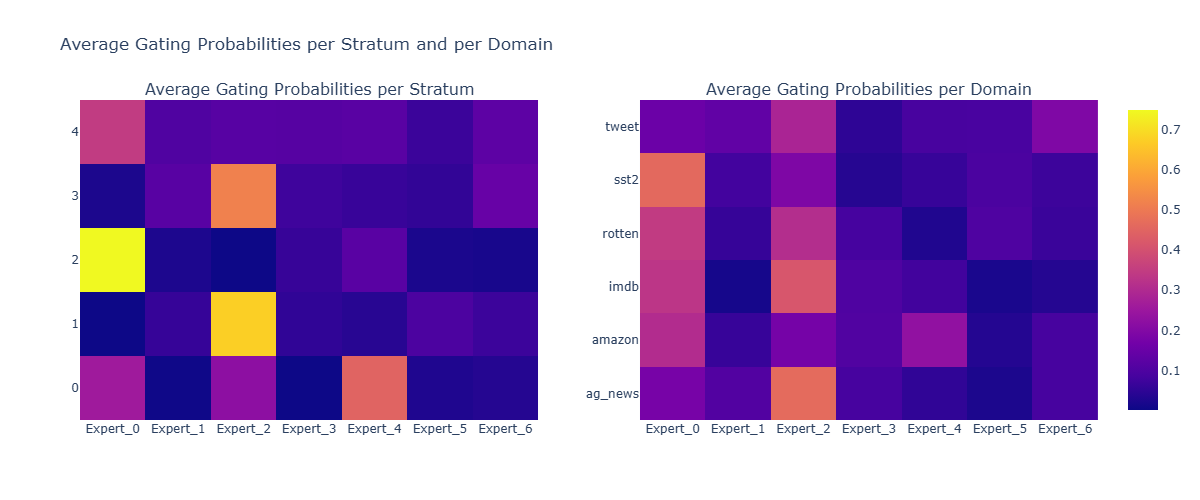}
    \caption{Gating probability visualizations for LLama-3.2-1B.}
    \label{fig:llama3-probs}
\end{figure*}

\begin{figure*}[hbt!]
    \centering
    \includegraphics[width=0.85\linewidth, height=12.5em]{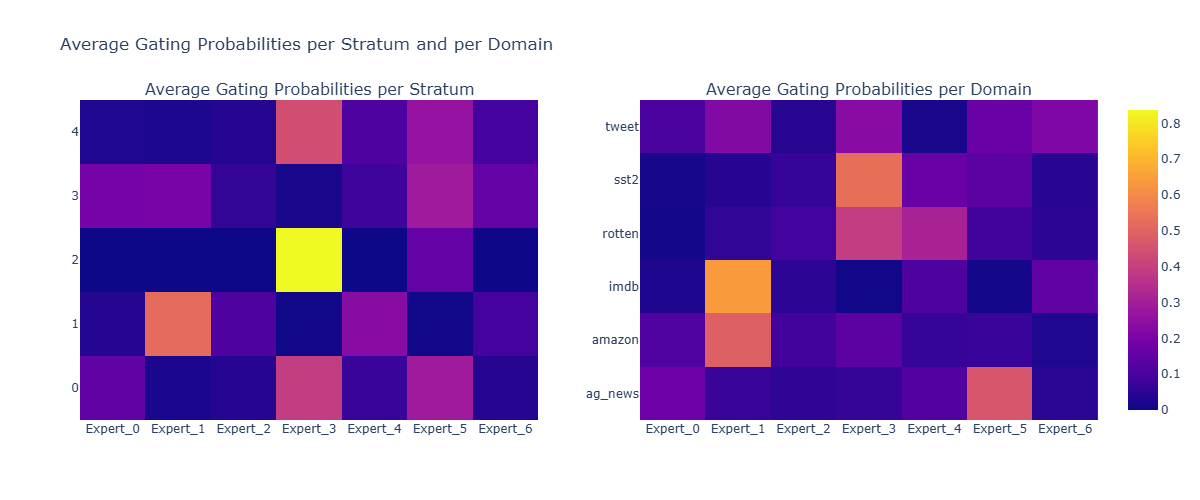}
    \caption{Gating probability visualizations for DeepSeek-R1-Distill-Qwen-1.5B.}
    \label{fig:deepseek-probs}
\end{figure*}

\section{Experiments}

To evaluate the stratification of the embedding space of large language models. We first create a mixture of text datasets from different domains, which includes the following,
\begin{description}
    \item[IMDB:] movie review dataset, which contains $25,000$ highly polar movie reviews for training and $25,000$ for testing, note that the movie reviews are rather long sentences, thus more complex \cite{imdb};
    \item[Rotten Tomatoes:] movie review dataset containing $5,331$ positive and $5,331$ negative processed sentences from Rotten Tomatoes moview reviews, which are of moderate lengths \cite{rotten-tomatoes};
    \item[Amazon Polarity Dataset:] dataset of reviews from Amazon, with a span of period of $18$ years and $\sim 35$ million reviews up to March $2013$, which includes information of products, users, ratings, reviews, which are of complex structures \cite{amazon-polarity};
    \item[GLUE/SST2:] the Stanford Sentiment Treebank dataset for the GLUE benchmark \cite{glue}, consisting of $11,855$ single sentences extracted from movie reviews, and parsed by the Stanford parser, including $215,154$ unique phrases from the parse trees, each annotated by $3$ human judges, we use this dataset as an instance of less complex dataset with simpler, shorter texts \cite{glue-sst2};
    \item[Tweets:] TweetEval dataset for sentiment analysis, with $45,615$ instances of labeled texts for training, $2,000$ for validation, and $12,284$ for testing, the data instances are essentially tweets and thus of moderate sizes \cite{tweet-sentiments};
    \item[AG News] news article dataset, consisting of more than $1$ million news articles from more than $2,000$ news sources in more than $1$ year of activity by an academic news search engine called \textit{ComeToMyHead}, which can be long and complex \cite{agnews}.
\end{description}

For our experiments \footnote{Please feel free to check our implementations and experiment results at our GitHub repository: \url{https://anonymous.4open.science/r/llm-stratified-934A}.}, we pick $500$ samples from each data source and merge them together to get a complex, multi-domain dataset to test the stratification of the LLM embedding space. As baseline LLM models, we have picked BERT \cite{bert}, RoBERTa \cite{roberta}, GPT-3 \cite{gpt}, LLama-3.2-1B \cite{llama3}, and DeepSeek-R1-Distill-Qwen-1.5B \cite{deepseek} based on our available hardware, of which we will present and analyze the results for RoBERTa, LLama-3.2-1B and DeepSeek-R1-Distill-Qwen-1.5B but we will also provide other results in Appendix \ref{subsec:appendix-exps}. For all of our experiments, we use the last hidden layer outputs as input embeddings, we have also set the number of dictionary atoms as $32$ with varying sparsity levels of $[8, 12, 16, 20, 24, 28, 32]$ for seven distinct dictionary learning experts, we also set the number of strata as $5$ for easier analysis of our results. We train our model on embeddings generated from the large language models we pick with a learning rate of $1 \times 10^{-3}$ for $100$ epochs on a single NVIDIA L4 Tensor Core GPU, the experiments run fairly quickly with around $2$ hours for a single experiment on the larger models (LLama-3.2-1B, DeepSeek-R1-Distill-Qwen-1.5B) and much quicker on smaller models like BERT.

\subsection{Visualizations}

Here we use PCA to project the high-dimensional embedding space to a $3D$ space to perform the clustering visualizations, and use stratum indices as labels, as mentioned in the previous sections, here we provide $3D$ clustering visualizations of learned stratum assignment in embedding space on RoBERTa, LLama-3 \cite{llama3}, and DeepSeek \cite{deepseek} models. We have also provided the visualizations of the gating probabilities regarding expert assignments and strata.

\begin{table*}[hbt!]
\centering
\begin{tabular}{lrrr}
\hline
 & \multicolumn{1}{l}{\textbf{RoBERTa}} & \multicolumn{1}{l}{\textbf{LLama-3.2-1B}} & \multicolumn{1}{l}{\textbf{DeepSeek-R1-Distill-Qwen-1.5B}} \\ \hline
\textbf{stratum 0} & 15 & 2  & 4  \\
\textbf{stratum 1} & 11 & 16 & 9  \\
\textbf{stratum 2} & 10 & 4  & 3  \\
\textbf{stratum 3} & 13 & 16 & 12 \\
\textbf{stratum 4} & 5  & 13 & 8  \\ \hline
\end{tabular}
\caption{Intrinsic stratum dimensions of LLM local embedding space.}
\label{tab:intrinsic-dims}
\end{table*}

\begin{table*}[hbt!]
\centering
\begin{tabular}{lrrr}
\hline
 & \multicolumn{1}{l}{\textbf{RoBERTa}} & \multicolumn{1}{l}{\textbf{LLama-3.2-1B}} & \multicolumn{1}{l}{\textbf{DeepSeek-R1-Distill-Qwen-1.5B}} \\ \hline
\textbf{stratum 0} & 20.6789 & 18.4351 & 20.9596 \\
\textbf{stratum 1} & 20.1984 & 18.5504 & 16.9576 \\
\textbf{stratum 2} & 19.7239 & 11.7021 & 21.2822 \\
\textbf{stratum 3} & 20.5041 & 19.1496 & 20.3541 \\
\textbf{stratum 4} & 19.5071 & 17.1080 & 22.8659 \\ \hline
\end{tabular}
\caption{Weighted sparsity levels of the dictionary learning experts assigned per stratum.}
\label{tab:weighted-sparsity}
\end{table*}

\subsection{Evaluation}

From our experiments, we can clearly see that embeddings from similar datasets are being clustered together and tend to share the same stratum, and we can observe that as compared an smaller model RoBERTa, bigger models like LLama-3.2-1B and DeepSeek-R1-Distill-Qwen-1.5B have a clearer separation of the embedding space regarding embeddings generated from different domains of inputs. We can also see from the gating probability visualizations bigger models display a more concentrated expert selection regarding the strata, which could imply the uniformity of the embedding space perplexity. For a smaller model like RoBERTa, the expert assignments are more spread out, which might mean the model's embedding space is not as smooth as the bigger models and thus require experts with different sparsity levels pretty uniformly. We can also see a slightly better concentration of expert assignments regarding the domains and strata in the DeepSeek-R1-Distill-Qwen-1.5B model than LLama-3.2-1B, which might mean the model generate smoother embedding space across strata and domains than the LLama-3.2-1B model.

To further analyze the stratification, we compute the intrinsic dimensions of different inferred strata using a simple PCA algorithm and trying to count the number of eigenvalues that can capture $75\%$ of the data variance, which are provided in Table \ref{tab:intrinsic-dims}. We have also computed the weighted sparsity level across different strata in Table \ref{tab:weighted-sparsity}. As we can see from the tables, the embedding space of these large language models do exhibit a non-uniform geometrical structure, with varying intrinsic dimensions and complexities (based on average sparsity levels) of different strata. This aligns with our hypothesis, which provides valuable insights in relevant research in the future.

\section{Conclusions}

In this paper, we have developed a framework to analyze the local geometrical structures of the LLM embedding space. In particular, we have developed a dictionary-learning based Mixture-of-Expert model with an attention-based gating mechanism to learn soft assignments of experts with different complexity parameters (sparsity levels in our case) to hypothetical strata in the space. Our results align with our hypothesis that the LLM embedding space do not have a smooth global manifold but rather a collection of local stratified manifolds of different dimensions, which provides valuable insights of fine-grained localized structure of the LLM embeddings, and opens the door to future research on smarter design choices and better understanding of the large language models or the like (this analysis can be easily adapted to the embedding spaces of other model architectures).

\section{Limitations}

Despite rather interesting experimental findings, we have yet investigated the mathematics details of this structure, and we also haven't worked on potential applications of this stratification idea as this is actually a fairly novel idea. Further work might include a more insightful and mathematical rigorous look into this structure and its properties, along with potential applications that can help improve LLM model performance, interpretability, or some other similar ideas.




\bibliography{main}

\appendix

\section{Appendix}
\label{sec:appendix}

\subsection{Review of Basic Mathematics Concepts}
\label{subsec:appendix-review}

Here we provide a more comprehensive view of the relevant concepts that can be helpful to understand the idea of stratification as discussed in the main body of the paper.

\subsubsection{Basics of Mathematical Analysis}

\begin{definition}[Equivalence Relation]
    If there exists a one-one mapping $f$ of a set $\mathsf{X}$ \textit{onto} another set $\mathsf{Y}$ ($f(\mathsf{X}) = \mathsf{Y}$, where the set $f(\mathsf{X)}$ denotes the range of function mapping $f$ on domain $\mathsf{X}$), we say that $\mathsf{X}$ and $\mathsf{Y}$, can be put in one-one \textit{correspondence}, or briefly, that $\mathsf{X}$ and $\mathsf{Y}$ are \textit{equivalent}, we denote $\mathsf{X} \sim \mathsf{Y}$. This relation is called the \textit{equivalence relation}, with the following properties,
    \begin{enumerate}
        \item $\mathsf{X} \sim \mathsf{X}$ (reflexivity);
        \item If $\mathsf{X} \sim \mathsf{Y}$, then $\mathsf{Y} \sim \mathsf{X}$ (symmetry);
        \item If $\mathsf{X} \sim \mathsf{Y}, \mathsf{Y} \sim \mathsf{Z}$, then $\mathsf{X} \sim \mathsf{Z}$ (transitivity).
    \end{enumerate}
\end{definition}

\begin{definition}[Countability]
    For any positive integer $n$, let $\mathsf{J}_n$ be the set whose elements are the integers $1, 2, \cdots, n$ and let $\mathsf{J}$ be the set consisting of all positive integers. For any set $\mathsf{X}$, we say:
    \begin{enumerate}
        \item $\mathsf{X}$ is \textit{finite} if $\mathsf{X} \sim \mathsf{J}_n$ for some $n$ (the empty set is also considered to be finite);
        \item $\mathsf{X}$ is \textit{infinite} if $\mathsf{X}$ is not finite;
        \item $\mathsf{X}$ is \textit{countable} if $\mathsf{X} \sim \mathsf{J}$;
        \item $\mathsf{X}$ is \textit{uncountable} if $\mathsf{X}$ is neither finite nor countable.
    \end{enumerate}
\end{definition}

\subsubsection{Basics of Topology}

\begin{definition}[Metric Spaces]
    A set $\mathsf{X}$, whose elements are called \textit{points}, is said to be a \textit{metric space} if for any two points $p$ and $q$ of $\mathsf{X}$, there is an associated real number $d(p, q)$, called the \textit{distance} from $p$ to $q$, such that
    \begin{enumerate}
        \item $d(p, q) \geq 0$ and $d(p, q) = 0 \iff p = q$;
        \item $d(p, q) = d(q, p)$ \textit{(symmetry)};
        \item $d(p, q) \leq d(p, r) + d(r, q)$, for any $r \in \mathsf{X}$ \textit{(triangle inequality)}.
    \end{enumerate}
    Any function with these properties is called a \textit{distance function}, or a \textit{metric}.
\end{definition}
One typical metric space is the Euclidean spaces $\mathsf{R}^n$, where $\mathsf{R}^1$ denotes the real line and $\mathsf{R}^2$ denotes the complex plane.

\begin{definition}[Ball]
    If $\boldsymbol{x} \in \mathsf{R}^k$ and $r > 0$, the \textit{open (or closed) ball} $\mathbf{B}$ centered at $\boldsymbol{x}$ with radius $r$ is defined to be the set of all $\boldsymbol{y} \in \mathsf{R}^k$ such that $|\boldsymbol{y} - \boldsymbol{x}| < r$ (or $|\boldsymbol{y} - \boldsymbol{x}| \leq r$).
\end{definition}

\begin{definition}[Neighborhood, Limit Point, Closed and Open Sets in Metric Spaces]
    Let $\mathsf{X}$ be a metric space, we have the following definitions of relevant concepts in metric spaces,
    \begin{enumerate}
        \item A \textit{neighborhood} of a point $p \in \mathsf{X}$ is a set $\mathsf{N}_r(p) \subset \mathsf{X}$ consisting of all points $q \in \mathsf{X}$ such that $d(p, q) < r$. The number $r$ is called the \textit{radius} of $\mathsf{N}_r (p)$;
        \item A point $p \in \mathsf{X}$ is a \textit{limit point} of the set $\mathsf{E}$ if every neighborhood of $p$ contains a point $q \neq p$ such that $q \in \mathsf{E}$;
        \item A set $\mathsf{E}$ is a \textit{closed set} if every limit point of $\mathsf{E}$ is a point of $\mathsf{E}$;
        \item A point $p$ is an \textit{interior point} of $\mathsf{E}$ if there is a neighborhood $\mathsf{N}$ of $p$ such that $\mathsf{N} \subset \mathsf{E}$;
        \item A set $\mathsf{E}$ is an \textit{open set} if every limit point of $\mathsf{E}$ is a point of $\mathsf{E}$.
        \item The \textit{closure} of a set $\mathsf{E}$ is the smallest closed set containing $\mathsf{E}$, written as $\mathsf{cl} (\mathsf{E})$, in other words, the intersection of all closed sets which contain $\mathsf{E}$.
    \end{enumerate}
\end{definition}

If we drop the dependence on the distance function from the neighborhood definition in metric spaces, we will have a more generic definition.
\begin{definition}[Neighborhoods]
    For a set $\mathsf{X}$ and for each point $x \in \mathsf{X}$, a nonempty collection of subsets of $\mathsf{X}$, called the \textit{neighborhoods} of $x$, which satisfy the following four axioms,
    \begin{enumerate}
        \item $x$ lies in each of its neighborhoods;
        \item The intersection of two neighborhoods of $x$ is itself a neighborhood of $x$;
        \item If $\mathsf{N}$ is a neighborhood of $x$ and if $\mathsf{U}$ is a subset of $\mathsf{X}$ which contains $\mathsf{N}$, then $\mathsf{U}$ is a neighborhood of $x$.
        \item If $\mathsf{N}$ is a neighborhood of $x$ and if $\mathsf{N}'$ denotes the set $\{z \in \mathsf{N} \mid \mathsf{N} \text{ is a neighborhood of } z \}$, then $\mathsf{N}'$ is a neighborhood of $x$, and we call the set $\mathsf{N}'$ the \textit{interior} of $\mathsf{N}$.
    \end{enumerate}
\end{definition}

Then we call $\mathsf{X}$ a \textit{topological space}, which is an extension on the metric spaces, where a subset $\mathsf{E} \in \mathsf{X}$ is called an \textit{open set} if it is a neighborhood of each of its points. And the assignment of a collection of neighborhoods that satisfy axioms 1 - 4 to each point $x \in \mathsf{X}$ is called a \textit{topology} on the set $\mathsf{X}$. Formally,

\begin{definition}[Topology]
    A \textit{topology} on a set $\mathsf{X}$ is a non-empty collection of subsets of $\mathsf{X}$, called open sets, such that any union of open sets is open, any finite intersection of open sets is open and both $\mathsf{X}$ and the empty set are open. A set together with a topology on it is called a \textit{topological space}.
\end{definition}

\begin{definition}[Open and Closed Sets]
    A subset of a topological space is \textit{open} is it is a neighborhood of each of its points, and is \textit{closed} if its complement is open.
\end{definition}

\begin{definition}[Limit Points in Topological Spaces]
    Let $\mathsf{E}$ be a subset of a topological space $\mathsf{X}$ and call a point $p \in \mathsf{X}$ a \textit{limit point} (or \textit{accumulation point}) of $\mathsf{E}$ if every neighborhood of $p$ contains at least one point of $\mathsf{E} - \{p\}$.
\end{definition}

A set is closed if and only if it contains all its limit points, and

\begin{definition}[Closure in Topological Spaces]
    The union of the subset $\mathsf{E}$ in a topological space $\mathsf{X}$ and all its limit points is called the \textit{closure} of $\mathsf{E}$ and is written as $\mathsf{cl} (\mathsf{E} )$.
\end{definition}
Note that the closure of the set $\mathsf{E}$ is the smallest closed set containing $\mathsf{E}$, in other words the intersection of all closed sets which contain $\mathsf{E}$.

\begin{definition}[Base for the Topology]
    Suppose we have a topology on a set $\mathsf{X}$, and a collection of open sets $\beta$ such that every open set is a union of members of $\beta$ is called a \textit{base} for the topology and the elements of $\beta$ are called \textit{basic open sets}.
\end{definition}

\subsubsection{Basics of Abstract and Commutative Algebra}

Now let's review a few concepts from abstract and commutative algebra,

\begin{definition}[Binary Operation]
    Let $\mathsf{G}$ be a set. A \textit{binary operation} $\cdot$ is a map of sets,
    \begin{equation}
        \ast : \mathsf{G} \times \mathsf{G} \mapsto \mathsf{G}.
    \end{equation}
    We can write $\ast (a, b) = a \ast b$ for all $a, b \in \mathsf{G}$.
\end{definition}
Intuitively, any binary operation on $\mathsf{G}$ gives a way of combining elements.

\begin{definition}[Group]
    A \textit{group} is a set $\mathsf{G}$, together with a binary operation $\cdot$, such that,
    \begin{enumerate}
        \item $(a \ast b) \ast c = a \ast (b \ast c)$ for all $a, b, c \in \mathsf{G}$ (associativity);
        \item There exists an element $1_\mathsf{G} \in \mathsf{G}$ such that $a \ast 1_\mathsf{G} = 1_\mathsf{G} \ast a = a$ for all $a \in \mathsf{G}$  (existence of identity);
        \item Given $a \in \mathsf{G}$, there exists $b \in \mathsf{G}$ such that $a \ast b = b \ast a = 1_\mathsf{G}$ (existence of inverses).
    \end{enumerate}
\end{definition}

A set with a binary operation is called a \textit{monoid} if only if the first two properties hold, a group is a monoid where every element is invertible.

\begin{definition}[Abelian Group]
    A group $(\mathsf{G}, \ast)$ is \textit{Abelian} if it also satisfies
    \begin{equation}
        a \ast b = b \ast a,\quad \forall a, b \in \mathsf{G},
    \end{equation}
    which is also called the \textit{commutative property}.,
\end{definition}

\begin{definition}[Ring]
    A \textit{ring} is a set $\mathsf{R}$ with two binary operations, $+$, called \textit{addition}, and $\times$, called \textit{multiplication}, such that
    \begin{enumerate}
        \item $\mathsf{R}$ is an Abelian group under addition;
        \item $\mathsf{R}$ is a monoid under multiplication, \textit{i.e.}, inverses do not necessarily exist;
        \item $+$ and $\times$ are related by the \textit{distributive law},
        \begin{align}
            (x + y) \times z &= x \times z + y \times z \nonumber\\
            x \times (y + z) &= x \times y + x \times z
        \end{align}
        for all $x, y, z \in \mathsf{R}$.
    \end{enumerate}
    Note that the identity for $+$ is ``zero'', denoted as $0_\mathsf{R}$ and the identity for $\times$ is ``one'', denoted as $1_\mathsf{R}$.
\end{definition}

\begin{definition}[Ideal]
    Let $\mathsf{R}$ be a ring. A subset $\mathsf{I}$ is called an \textit{ideal} if
    \begin{enumerate}
        \item $0_\mathsf{R} \in \mathsf{I}$;
        \item $x + y \in \mathsf{R}$ for all $x, y \in \mathsf{R}$;
        \item $x e \in \mathsf{I}$ for all $x \in \mathsf{R}$ and $e \in \mathsf{I}$.
    \end{enumerate}
\end{definition}

\begin{definition}[Module]
    Let $\mathsf{R}$ be a ring, an $\mathsf{R}$-\textit{module} $\mathsf{M}$ is an Abelian group, written additively, with a scalar multiplication $\mathsf{R} \times \mathsf{M} \mapsto \mathsf{M}$, denoted $(x, m) \mapsto xm$, such that
    \begin{enumerate}
        \item $x (m + n) = xm + xn$ and $(x + y)m = xm + xm$ (distributivity);
        \item $x (ym) = (xy) m$ (associativity);
        \item $1_\mathsf{R} \cdot m = m$ (identity).
    \end{enumerate}
    A \textit{submodule} $\mathsf{N}$ of $\mathsf{M}$ is a subgroup that is closed under multiplication, \textit{i.e.}, $xn \in \mathsf{N}$ for all $x \in \mathsf{R}$ and $n \in \mathsf{N}$.
\end{definition}

\begin{definition}[Filtration]
    Let $\mathsf{R}$ be an arbitrary ring and $\mathsf{M}$ a module. A \textit{filtration}  $\mathcal{F}$ of $\mathsf{M}$ is an indefinite ascending chain of submodules,
    \begin{equation}
        \mathsf{M} \subset \cdots \subset \mathcal{F}^n \mathsf{M} \subset \mathcal{F}^{n + 1} \mathsf{M} \subset \cdots,
    \end{equation}
\end{definition}

\subsubsection{Smoothness of Manifolds}

Finally, to understand the smoothness of a manifold, we provide the following concept defintions,

\begin{definition}[Chart]
    A \textit{chart} is a pair $(\mathsf{U}, \phi)$ where $\mathsf{U}$ is an open set in $\mathsf{X}$ and $\phi: \mathsf{U} \mapsto \mathsf{R}^n$ is a homeomorphism. Given two charts $(\mathsf{U}_1, \phi_1)$ and $(\mathsf{U}_2, \phi_2)$, we can define the \textit{overlap/transition maps},
    \begin{align}
        \phi_2 \circ \phi_1^{-1} &: \phi_1 (\mathsf{U}_1 \cap \mathsf{U}_2) \mapsto \phi_2 (\mathsf{U}_1 \cap \mathsf{U}_2), \nonumber\\
        \phi_1 \circ \phi_2^{-1} &: \phi_2 (\mathsf{U}_1 \cap \mathsf{U}_2) \mapsto \phi_1 (\mathsf{U}_1 \cap \mathsf{U}_2).
    \end{align}
\end{definition}

\begin{definition}[Compatible Charts]
    Two charts $(\mathsf{U}_1, \phi_1)$ and $(\mathsf{U}_2, \phi_2)$ are \textit{compatible} if the overlap maps are smooth.
\end{definition}

\begin{definition}[Atlas]
    An \textit{atlas} for $\mathsf{X}$ is a non-redundant collection $\mathsf{A} = \{ (\mathsf{U}_\alpha, \phi_\alpha) | \alpha \in \mathsf{A} \}$ of pairwise compatible charts.
\end{definition}

\begin{definition}[Smooth Manifold]
    A smooth $n$-dimensional manifold is a Hausdorff, second countable, and topological space $\mathsf{X}$ (manifold) together with an atlas $\mathsf{A}$.
\end{definition}

\subsection{More Results and Visualizations}
\label{subsec:appendix-exps}

\begin{figure*}[hbt!]
    \centering
    \includegraphics[width=\linewidth]{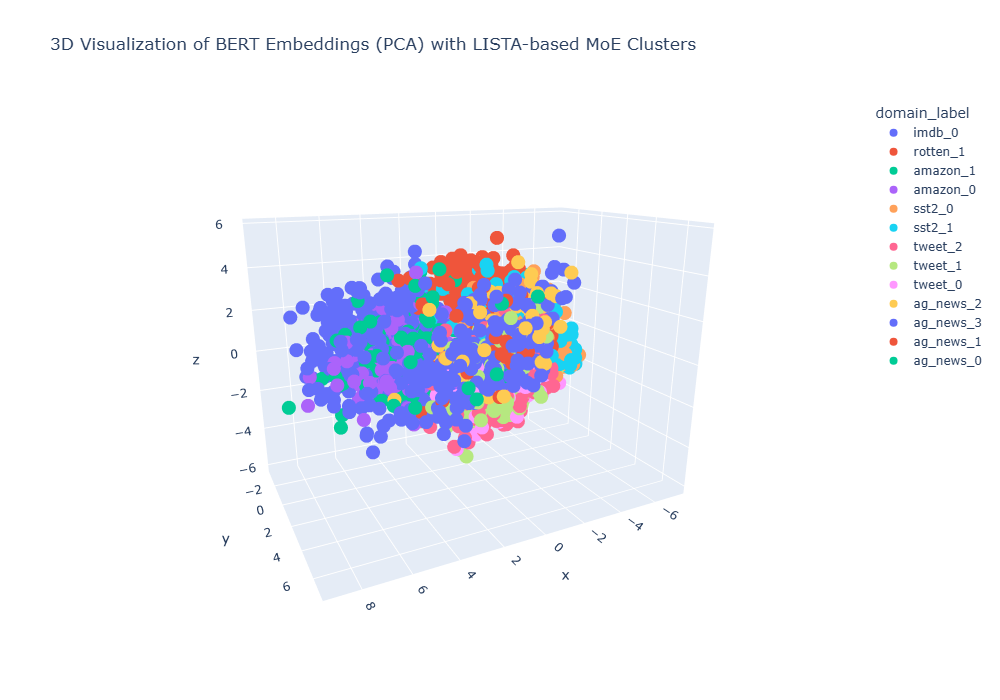}
    \label{fig:bert}
\end{figure*}

\begin{figure*}[hbt!]
    \centering
    \includegraphics[width=\linewidth]{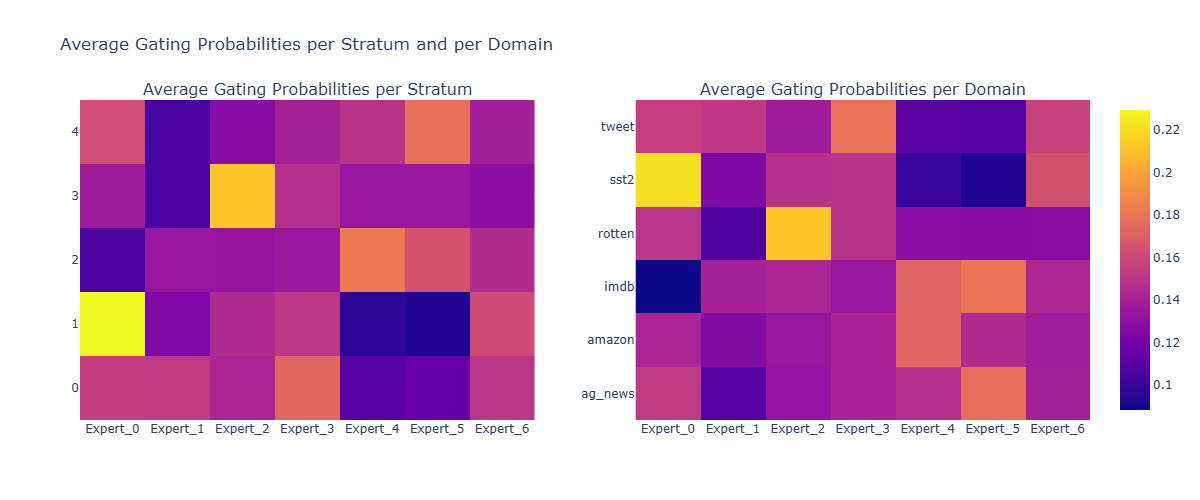}
    \caption{Gating probability visualizations for BERT.}
    \label{fig:bert-probs}
\end{figure*}

\begin{figure*}[hbt!]
    \centering
    \includegraphics[width=\linewidth]{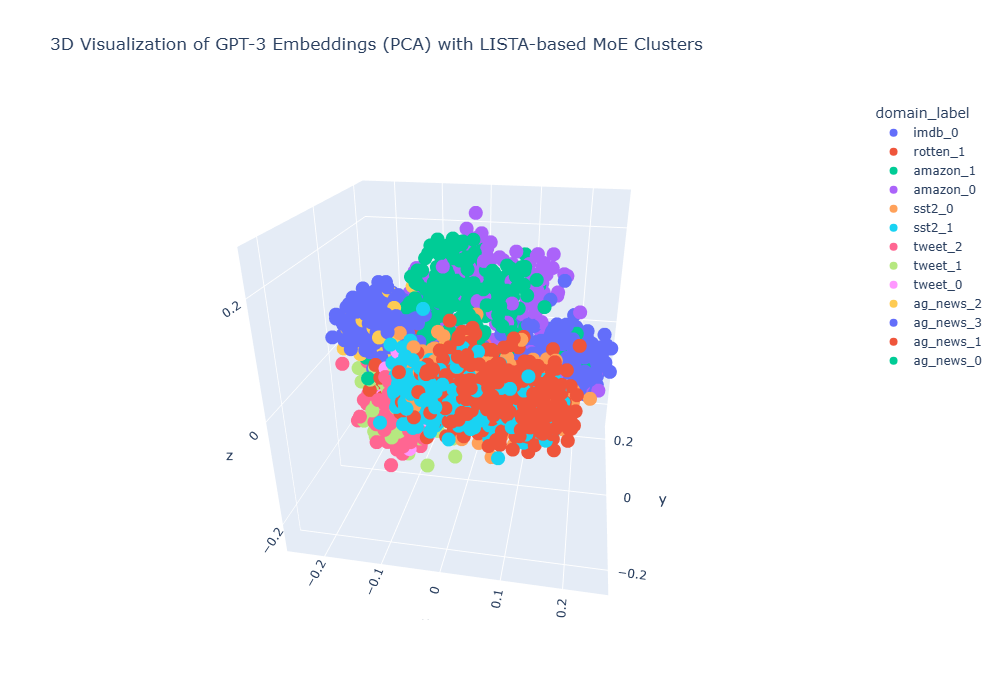}
    \label{fig:gpt}
\end{figure*}

\begin{figure*}[hbt!]
    \centering
    \includegraphics[width=\linewidth]{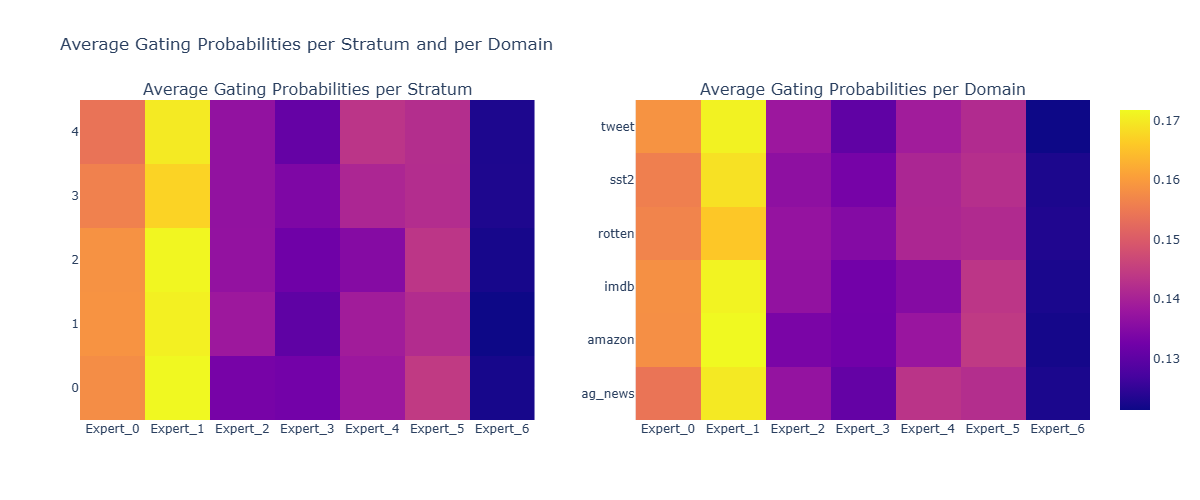}
    \caption{Gating probability visualizations for GPT-3.}
    \label{fig:gpt3-probs}
\end{figure*}

\end{document}